\documentclass[fleqn,10pt]{wlscirep}
\usepackage[utf8]{inputenc}
\usepackage[T1]{fontenc}

\usepackage{amsmath}
\usepackage{tabularx}
\usepackage{multirow}

\usepackage{xurl}

\usepackage{caption}
\usepackage{subcaption}
\usepackage{float}
\floatstyle{plain}
\restylefloat{figure}

\newcolumntype{Y}{>{\centering\arraybackslash}X}
\newcommand{\etal}{\textit{et al.}}

\usepackage{nameref}
\usepackage{hyperref}
\usepackage[capitalise, noabbrev]{cleveref}

\title{Satellite-based high-resolution maps of cocoa planted area for Côte d’Ivoire and Ghana}

\author[1,*]{Nikolai Kalischek}
\author[1]{Nico Lang}
\author[2]{C\'{e}cile Renier}
\author[1]{Rodrigo Caye Daudt}
\author[3]{Thomas Addoah}
\author[4]{William Thompson}
\author[5]{Wilma J. Blaser-Hart}
\author[3]{Rachael Garrett}
\author[1]{Konrad Schindler}
\author[1,6]{Jan D. Wegner}
\affil[1]{EcoVision Lab, Photogrammetry and Remote Sensing, ETH Zurich, 8092, Zurich, Switzerland}
\affil[2]{Earth and Life Institute, UC Louvain, 1348 Louvain-la-Neuve, Belgium}
\affil[3]{Environmental Policy Lab, Department of Humanities, Social and Political Sciences, ETH Zurich, 8092, Zurich, Switzerland}
\affil[4]{Department of Biology, University of Oxford, Oxford, OX1 2JD, Great Britain}
\affil[5]{School of Biological Sciences, The University of Queensland, St Lucia, Brisbane, QLD 4072, Australia}
\affil[6]{Data Science for Sciences, Institute for Computational Science, University of Zurich, 8092, Zurich, Switzerland}

\affil[*]{nikolai.kalischek@geod.baug.ethz.ch}

\begin{abstract}
Côte d’Ivoire and Ghana, the world’s largest producers of cocoa, account for two thirds of the global cocoa production. In both countries, cocoa is the primary perennial crop, providing income to almost two million farmers. Yet precise maps of cocoa planted area are missing, hindering accurate quantification of expansion in protected areas, production and yields, and limiting information available for improved sustainability governance. Here, we combine cocoa plantation data with publicly available satellite imagery in a deep learning framework and create high-resolution maps of cocoa plantations for both countries, validated in situ. Our results suggest that cocoa cultivation is an underlying driver of over 37 \% and 13 \% of forest loss in protected areas in Côte d’Ivoire and Ghana, respectively, and that official reports substantially underestimate the planted area, up to 40 \% in Ghana. These maps serve as a crucial building block to advance understanding of conservation and economic development in cocoa producing regions.

\end{abstract}
\begin{document}

\flushbottom
\maketitle


\thispagestyle{empty}


Cocoa is grown by an estimated 2 million farmers in West Africa, who supply a complex network of middle men, private and public companies, which renders the supply chain rather opaque \cite{schulte2020supporting, carodenuto2021effect, zu2022addressing}. With an average farm size of three to five hectares \cite{hainmueller2011sustainable, bymolt2018demystifying} and an estimated income of less than one dollar per day, nearly all cocoa farmers live under the poverty line \cite{earth2017chocolate}. In this context, deforestation in West African Upper Guinean forests, a biodiversity hotspot \cite{cepf_biodiversity}, has occurred in waves across the 20\textsuperscript{th} and 21\textsuperscript{st} centuries \cite{fairhead2003reframing}. Cocoa-driven deforestation has played a significant role in this and has been catalyzed by migration from Savannah regions, land availability and tenure constraints for residents of existing cocoa production areas, as well as the higher productive potential of recently cleared land \cite{schulte2020supporting, ruf2015climate}.

In recent years, corporate sustainability efforts have been initiated to reduce cocoa-driven deforestation, improve cocoa yields, including by promoting agroforestry \cite{worldcocoafoundation2021}. While cocoa certification programs have improved farm productivity and income, there is no conclusive evidence of their impact on agroforestry or deforestation. Furthermore, more ambitious supply chain targets to eliminate sourcing of deforestation-linked cocoa face major implementation challenges due to difficulty in monitoring and tracking cocoa expansion into forests \cite{lambin2018role, mightyearth2022}. Among mining, selective logging and other crops, cocoa has been the primary driver of deforestation in these countries \cite{ruf2015climate, barima2016cocoa, carodenuto2021effect}. However the extent to which cocoa has directly and indirectly replaced forest, until now, is uncertain.

Current map products are either derived from small reference data sets and of low precision, or rely on manual geo-referencing and are costly to update \cite{mightyearth2022, abu2021detecting}. The production of accurate, high-resolution maps of cocoa growing areas is currently missing. Up-to-date maps could greatly enhance efforts to halt deforestation by highlighting high-deforestation risk sourcing areas for cocoa, for verifying production quantities, and estimating on-farm versus off-reserve production area. Beyond deforestation, the spatial extent of cocoa production could be linked with more readily available data on production quantities to inform more targeted extension activities. 

Here, we present a large-scale, high-resolution cocoa growing map spanning Côte d'Ivoire and Ghana, generated by satellite image analysis with a deep neural network. Deep learning has matured and surpassed traditional hand-crafted feature detectors in countless remote sensing tasks including vegetation height mapping \cite{lang2019country}, localizing fires \cite{barmpoutis2020review}, predicting photovoltaic solar facilities \cite{kruitwagen2021global} and crop identification \cite{rodriguez2021mapping, turkoglu2021crop}. When trained on large reference corpora, deep models offer an unprecedented ability to recognize visual patterns in unseen data. 

\begin{figure*}[!bt]
    \centering
    \begin{subfigure}[b]{\textwidth}
        \includegraphics[width=1.0\textwidth, page=1]{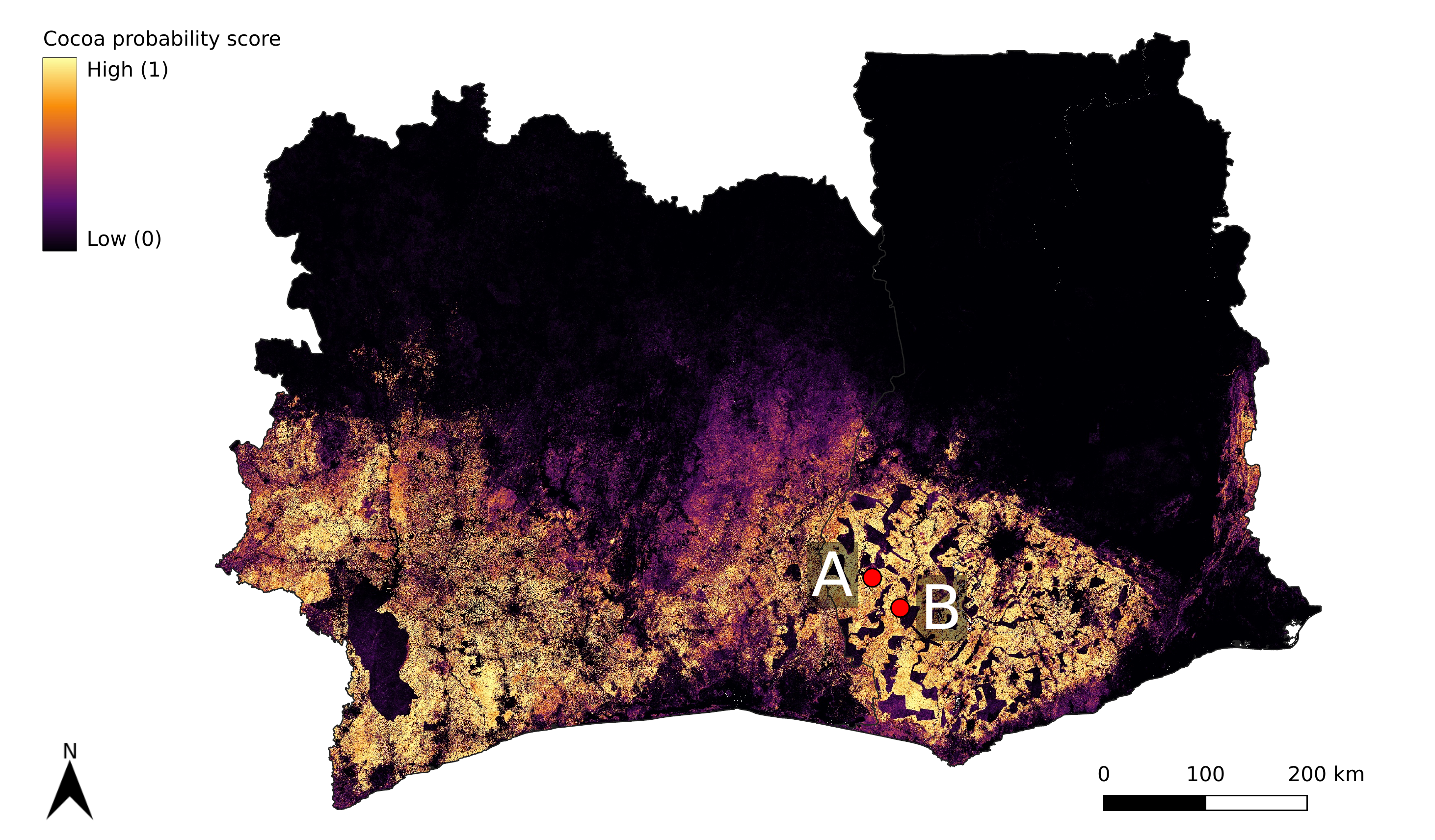}
    \end{subfigure}
    \begin{subfigure}[b]{\textwidth}
        \includegraphics[width=\textwidth, page=4]{images/cocoa_paper_1.pdf}
    \end{subfigure}
    \caption{\textbf{Cocoa map for Côte d'Ivoire and Ghana.} Probability map with 10$\times$10$\,$m ground sampling distance. The map indicates detection confidence on a range $[0\hdots 1]$, i.e., values near 1 indicate that model predictions across most time steps agree on the presence of cocoa, values near 0 agree on the absence of cocoa. \textbf{(A and B)} Two forested regions in Ghana where our model has detected cocoa farming (satellite images and confidence maps). The location of these forest regions are shown with red dots in the cocoa map.}\label{fig:confidence_map}
\end{figure*}

For this work, we have trained a neural network on a dataset of \textgreater100,000 geo-referenced cocoa farms to map cocoa plantations at country scale. We leverage publicly available optical satellite imagery as input in a twofold way. First, we train a neural network to predict canopy height in the sub-Saharan region utilizing ground truth acquired from the GEDI mission \cite{dubayah2020global}. Second, we train a deep neural network on the same satellite imagery and a large corpus of polygons delineating cocoa farms, using the canopy height map as an additional input for the network, introducing an explicit prior on the plant height. With the help of a team in Côte d'Ivoire, we validate our map \textit{in situ} in a three month long campaign, accompanied by further verification with a partially hand-labelled test set in Ghana. Instead of a single binary cocoa map, we create a probability map by aggregating predictions from a model ensemble and from repeated observations of the same location, as model ensembles have been shown to yield more reliable uncertainty calibration, and often also improved predictive skill \cite{lakshminarayanan2017simple}. The final map, along side two examples of cocoa grown in different protected areas, is depicted in \cref{fig:confidence_map}.

We illustrate the utility of our map, not only by (1) analyzing planted area, but also on (2) farming practices and sustainability efforts on reducing deforestation, highlighting the need for land cover mapping independent of farmers, industry and governments, and (3) identify regional areas that are exposed to poor growing conditions. 

\section*{Results}

\subsection*{Evaluation}
\begin{figure}[!tb]
    \centering
    \begin{subfigure}{\textwidth}
        \caption{}\label{fig:predictions}
        \includegraphics[width=\textwidth, page=5]{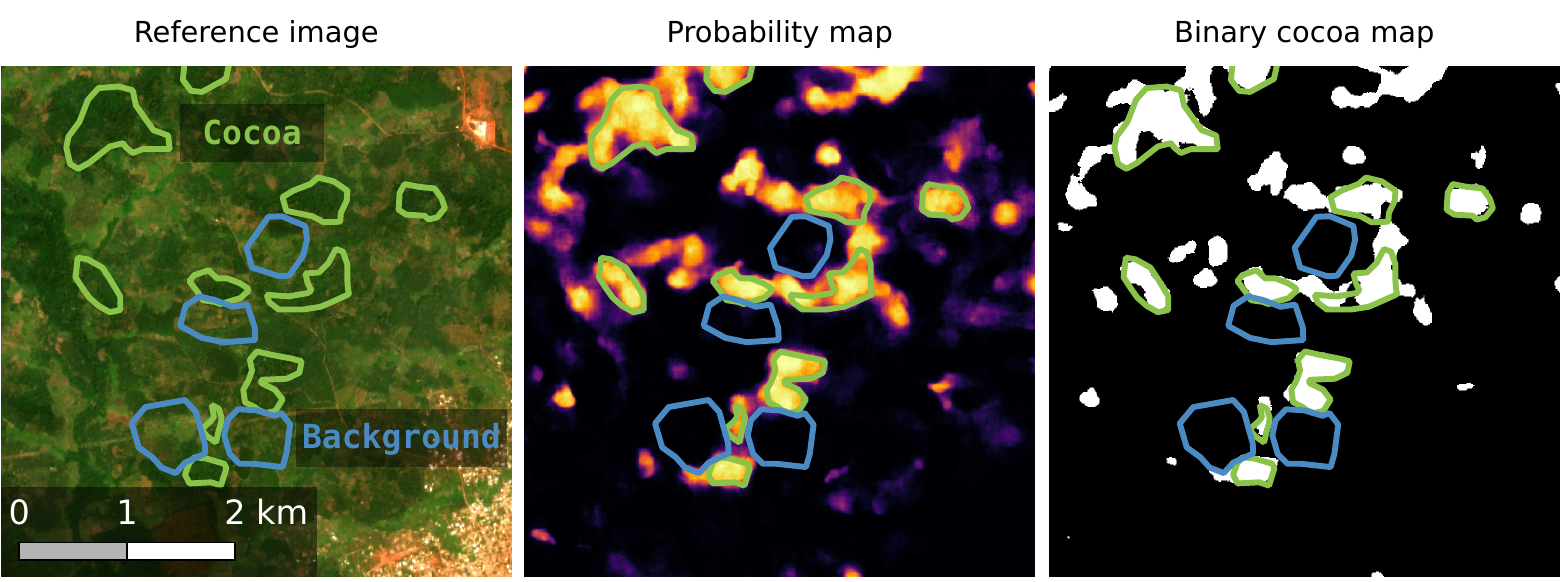}
    \end{subfigure}
    \begin{subfigure}{\textwidth}
        \caption{}\label{tab:metrics}
        \def\arraystretch{1.4}
        \begin{tabularx}{1\textwidth}{l|Y|Y|Y|Y}
        \hline
        Class & Precision (\%) & Recall (\%) & F1 (\%) & Accuracy (\%)\\
        \hline\hline
        Cocoa & 88.5 & 87.2 & 87.3 & \multirow{2}{*}{85.9} \\
        \cline{1-4}
        Non-cocoa & 83.9 & 84.3 & 84.1 \\
        \hline
        \end{tabularx}
    \end{subfigure}
    \caption{\textbf{In situ evaluation.} (a) Sites from our \textit{in situ} test set around Divo, Côte d'Ivoire. Left to right: a satellite image with reference data, confidence predictions and binary cocoa maps at a confidence threshold of 0.65. The green sites are mapped cocoa farms, the blue sites verified non-cocoa sites. In the binary map each value is either 0 or 1, i.e. each area is classified as either non-cocoa (0) or cocoa (1), whereas in the probability map values can be in between. (b) Quantitative performance of cocoa detection (confidence threshold 0.65). Contains modified Copernicus Sentinel data [2020].}
\end{figure}

First, we demonstrate the reliability of our ensemble model with four standard accuracy metrics. Precision (user's accuracy) measures the proportion of correctly classified pixels among all pixels assigned to a class. Recall (producer's accuracy) is the proportion of correctly classified pixels among all pixels that truly belong to a class. As summary statistics, we additionally report accuracy, i.e. overall fraction of correctly classified pixels, and $F_1$-score, defined as the harmonic mean of precision and recall for a class $c$:
\begin{equation*}
    F_1^c = 2 \cdot \frac{\text{recall} \cdot \text{precision}}{\text{recall} + \text{precision}}.
\end{equation*}
To avoid evaluation bias \cite{powers2020evaluation}, we present all metrics for both classes, cocoa and background, in \cref{tab:metrics} and showcase one examples of our \textit{in situ} test set in \cref{fig:predictions}.

A confidence map brings several benefits, one notable advantage is that one can calibrate the optimal threshold for binary cocoa presence/absence mapping with an additional, much smaller validation set for specific regions or applications.
We do this in order to evaluate our cocoa map against two independent, binary validation datasets. In Côte d'Ivoire, we manually labeled over 2,000 polygons and verified these estimates with on-farm visits. Additional information on this \textit{in situ} dataset is given in Methods. For Ghana, we acquire geo-referenced cocoa polygons from an independent commercial data provider. While our in situ test set covers various regions in both countries, it is restricted to the areas around major cities in Côte d'Ivoire, since it was infeasible to collect a statistically rigorous (stratified) random sample \cite{olofsson2014good} over a region of this scale.

Compared to previous mapping efforts, our approach offers several advantages. First, by using an end-to-end trainable framework, feature selection is automated. 
Our approach boosts all metrics by large margins compared to the only other large-scale cocoa map we are aware of \cite{abu2021detecting}, improving precision and recall by more than 30~\% and 8~\%, respectively. In terms of mapping effort, existing accountability maps \cite{cocoa_accountability_map_2021, interactive_cocoa_farm_map} rely on extensive collaborations with cocoa cooperatives to create, update and maintain databases of cocoa farms, which impedes their extension to underrepresented regions, both within a country and beyond. In contrast, our mapping system is naturally expandable to areas that have not previously been mapped. Our mapping system can be adjusted with a small amount of local reference data when expanded to a new area that enclose similar plant characteristics. For example, our map detects cocoa plantations in regions that, so far, have been ignored in official figures \cite{ghana_national_land_use_map}, e.g. the Volta region in Ghana.

\subsection*{Planted area}
With our cocoa map, we are able to calculate the total planted area in Côte d'Ivoire and in Ghana, and to compare it to official figures. We compute the best threshold at 0.65 according to a held-out validation set, i.e. all values above 65~\% in the probability map are classified as cocoa and all others as non-cocoa areas. Importantly, the probabilistic mapping approach makes it possible to select the threshold that maximizes the F1 score over the validation set, thus balancing the expected precision and recall on unseen data.We empirically found maximizing the F1-score to be more robust than direct matching of 
false negative and false positive counts: under practical conditions, where the training and test data are not perfect random samples from the underlying distribution, it leads to a lower area bias.The corresponding curve, where F1 score is plotted against different thresholds, can be found in the supplementary information.

To estimate planting areas, the uncertainty quantification via our model ensemble approach has several advantages. It has been argued that plain pixel counting in a binary classification map can be problematic for area estimation \cite{olofsson2014good}. However, our final map consists of continuous probability values, similar to model-based area estimation \cite{mcroberts2006model}, that can be thresholded in a post-processing step to minimize the bias of the area estimates. The theoretical scale factor between the true and estimated areas is $\frac{\text{precision}}{\text{recall}}$. For our test results that factor amounts to 1.02, i.e., a difference of less than 2\%. Moreover, each ensemble member produces its own (continuous) cocoa map and can be thresholded separately, thus drawing multiple samples from the distribution of area estimates. Contrary to naive pixel counting, we can thus characterize the uncertainty of the planted area estimate, by computing a mean area with an associated standard deviation and confidence intervals, assuming an underlying $t$-distribution.
The alternative, computing area estimates via optimally selected, stratified samples \cite{olofsson2014good, stehman2019key} is difficult for large-scale mapping efforts, particularly when the map resolution is high (in our case 10 m). To obtain such samples one would either need access to images of even higher resolution anywhere in the region of interest, so as to perform photo-interpretation. Or one must collect in situ data across entire countries, which is often not possible due to the difficulties of accessing randomly sampled locations that may lie in difficult terrain, lack infrastructure, be subject to land rights, etc.

In 2021, the mean total area under cultivation amounts to 4.45~Mha (CI 95~\%: 3.95 -- 4.95~Mha) in Côte d'Ivoire and 2.71~Mha (CI 95~\%: 2.21 -- 2.89~Mha) in Ghana, corresponding to 13.8~\% and 11.4~\% of Côte d'Ivoire's and Ghana's land area respectively. The detected cocoa plantings align well with climatically suitable growing regions in both countries \cite{laderach2013predicting}, although we have not restricted the detection to those areas \textit{a priori}, as in previous mapping projects \cite{abu2021detecting}.
Compared to the official FAOSTAT figures \cite{faostat}, our result deviates only marginally from the harvested area (average 2017--2020) in Côte d'Ivoire, but drastically differs for Ghana's total harvested area. FAOSTAT reports 4.47~Mha harvested area in Côte d'Ivoire (i.e. 0.5~\% more than our estimate) and only 1.63~Mha in Ghana, i.e. 39.8~\% less than our estimate. 

While the country-wide harvesting numbers are impressive by themselves for a single agricultural commodity, zooming in on a regional level further reveals the massive impact of cocoa cultivation on the two countries. As shown in \cref{fig:planted_area_per_region}, the highest proportions for a single region are 43.0~\% (Bas-Sassandra CI 95~\%: 1.16 -- 1.22~Mha) and 44.6~\% (Western CI 95~\%: 1.06 -- 1.11~Mha) for Côte d'Ivoire and Ghana, respectively, leaving little to no forested area in agricultural regions. This reveals the extent to which cocoa production has replaced native forests, which is associated with biodiversity loss, local and global climate impacts and the loss of multiple resources supporting food security and livelihoods.

\subsection*{Production}
\label{sec:production}

\begin{figure}[!tb]
    \centering
    \begin{subfigure}[b]{0.49\textwidth}
        \centering
        \caption{}\label{fig:planted_area_per_region}
        \includegraphics[width=\textwidth]{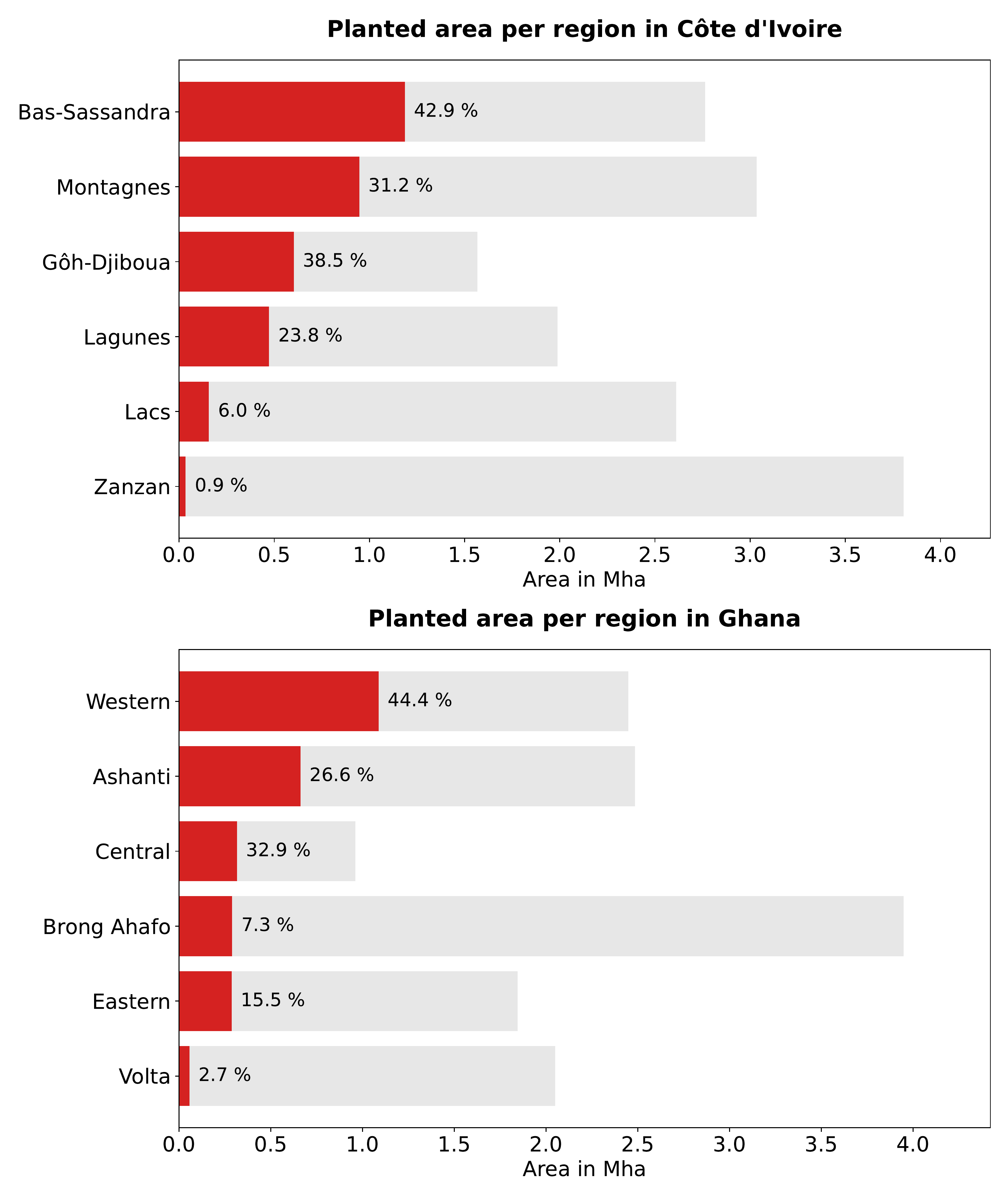}
    \end{subfigure}
    \begin{subfigure}[b]{0.49\textwidth}
        \centering
        \caption{}\label{fig:production:mean}
        \includegraphics[width=\textwidth, page=2]{images/cocoa_paper_1.pdf}
    \end{subfigure}
    \caption{\textbf{Planted area, mean annual yield and correlation between growing area and production volume per region.} \textbf{a}, Comparison of planted area and total area per region. Top figure depicts regions in Côte d'Ivoire, bottom figure in Ghana. The grey bars represent the total area of a region, while the red shaded bars show the absolute cocoa planted area. The percentages indicate the relative amount of cocoa planted area to the total area per region. \textbf{b}, Regional yield differences measured in kilogram per hectare in Ghana. In case of the Volta region, we only obtained production data for a single subdivision, hence the low average yield. }\label{fig:production}
\end{figure}

Using production data obtained from the Ghana Cocoa Board (COCOBOD) \cite{cocobod} and averaged over the years 2017 to 2020, we compare planted area with production data on a regional level in Ghana (production data at subnational level was not available for Côte d'Ivoire). By dividing production by planted area in every region, we obtain local mean yield estimates, important for farming practices, sustainability and regeneration efforts. Mean annual yields range from 250 kg/ha in Ashanti region to over 380 kg/ha in the Central region. The mean yields for the Volta region (120 kg/ha) need to be treated with care in our analysis, because production is reported only for one out of four cocoa growing districts.
In \cref{fig:production:mean}, we highlight significant differences in farming productivity within the cocoa growing regions in Ghana. As an example, even though the Western region contributes \textgreater40\% of the total growing area within the COCOBOD boundaries, the average yield is lower compared to the Central region, suggesting potential to improve farming practices. Even more extreme is the Ashanti region, where the annual yield is as low as 250 kg/ha. On average, we obtain a mean annual yield of 320 kg/ha for Ghana. Earlier studies have reported average cocoa yields in the range of 400 to 530 kg/ha \cite{lambert2014fairness, donovan2016fairtrade, vigneri2016researching, bymolt2018demystifying}, which is significantly higher compared to our estimate. These earlier numbers are based on small sets of field samples, which may be biased towards farms with above-average productivity \cite{bymolt2018demystifying}. Our yield estimates may also be slightly lower due to young planted farms detected by our map that may not be productive yet, therefore bringing down the average yield per unit area.  

\subsection*{Protected areas}

\begin{figure}[!tb]
    \centering
    \begin{subfigure}{\textwidth}
        \small
        \caption{}\label{tab:protected_areas}
        \begin{tabularx}{1\textwidth}{p{2.5cm}|Y|Y|Y||p{2.5cm}|Y|Y|Y}
        \hline
        \multicolumn{4}{l||}{Côte d'Ivoire} & \multicolumn{4}{l}{Ghana}\\
        \hline
        Protected area & Cocoa (ha) & Land cover (\%) & Deforestation (\%) & Protected area & Cocoa (ha) & Land cover (\%) & Deforestation (\%)\\
        \hline\hline
        Niegre (CF) & 108,256 & 81.8 & 86.6 & Tano Ehuro (FR) & 16,275 & 77.6 & 82.2\\
        Scio (CF) & 90,418 & 68.2 & 78.3 & Manzan (FR) & 15,512 & 56.1 & 48.6\\
        Mt. Sassandra (CF) & 54,946 & 49.0 & 55.8 & Upper Wassaw (FR) & 3,198 & 23.6 & 14.6 \\
        Mt. Péko (NP) & 6,479 & 21.5 & 19.3 & Sui River (FR) & 3,497 & 9.84 & 39.3 \\
        Marahoué (NP) & 2,789 & 18.0 & 13.2 & Kakum (NP) & 256 & 1.0 & 21.0\\
        \hline
        \end{tabularx}
        \vspace{3mm}
    \end{subfigure}
    \begin{subfigure}{\textwidth}
        \caption{}\label{fig:protected_areas_v2}
        \includegraphics[width=\textwidth,page=7]{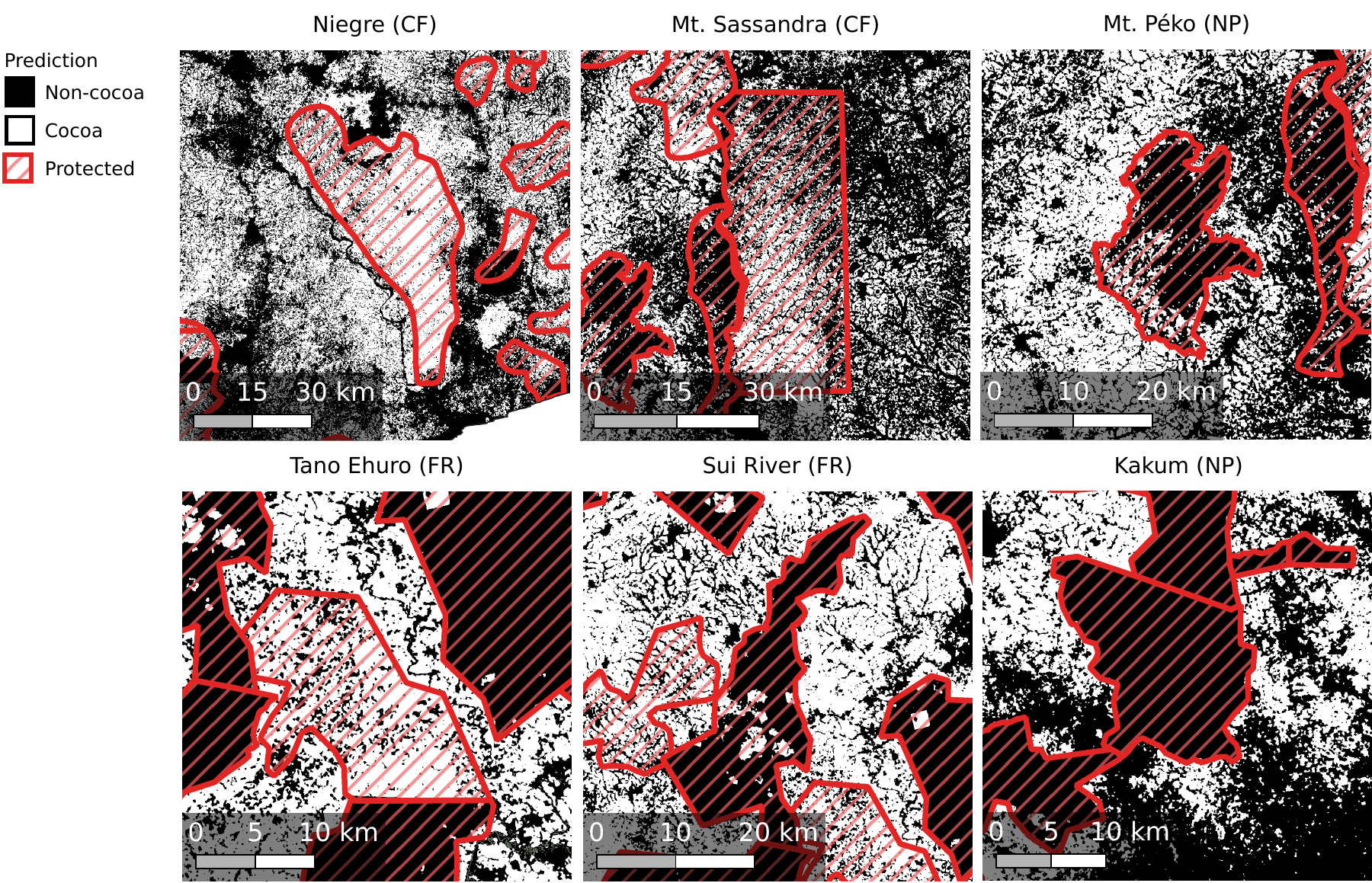}
    \end{subfigure}
    \caption{\textbf{Cocoa encroachment into protected areas.} (a) List of selected protected areas. Land cover is the proportion of cocoa within the protected area. Deforestation indicates the proportion of cocoa grown on deforested areas. CF is short for classified forest, FR is short for forest reserve and NP short for national park. Forest reserves are categorized as protected areas with sustainable use of natural resources. Full list can be found in \cref{tab:protected_areas_extended} in the Appendix. (b)  Maps of selected protected areas.}
    \label{fig:my_label}
\end{figure}

Côte d'Ivoire and Ghana continue to experience high forest loss. Côte d'Ivoire is estimated to have lost more than 90~\% of its forest cover since 1950, while Ghana has incurred forest losses of more than 65~\% \cite{mightyearth2022}. Forest clearance rates reached a high in 2018, increasing by 60~\% over 2017 in Ghana and by 26~\% in Côte d'Ivoire, the two highest increases in annual deforestation rates worldwide \cite{weisse2019world}. Though deforestation rates have fluctuated since 2018. 

Here, we examine to what extent cocoa replaced native forest. Our map enables an accurate assessment of cocoa related deforestation within protected areas mapped in the World Database on Protected Areas (WDPA) \cite{protectedplanet}. The WDPA includes various types of protected areas, including strict nature reserves, national parks and protected landscapes. Across all management categories, there are 242 and 286 protected areas in Côte d'Ivoire and in Ghana, respectively. Across these areas, we find a total cocoa planted area of \textgreater1.5~Mha located in protected areas: 1.3~Mha in Côte d'Ivoire, corresponding to 30~\% of the total cocoa area of the country, and 0.2~Mha in Ghana (i.e. 7~\% of the total cocoa area). These numbers correspond to \textgreater13.6~\% of the overall protected area in Côte d'Ivoire (9.8~Mha), and \textgreater4.5~\% in Ghana (3.7~Mha). Using the annually updated forest cover loss \cite{hansen2013high}, we can directly relate forest loss of over 360,000 hectares in protected areas (including classified forests) to cocoa cultivation in Côte d'Ivoire from 2000 to 2020. Given an overall forest loss of 962,000 hectares since 2000, cocoa is directly or indirectly responsible for almost 37.4~\% of forest loss in protected areas. Similarly, we can trace back 26,000 hectares of cocoa driven deforestation in protected areas in Ghana, corresponding to 13.5~\% of the total forest lost in protected areas (193,000 ha) since 2000.

We further break down the numbers to individual protected areas. \cref{tab:protected_areas} lists the total cocoa area and the corresponding relative land cover and deforestation percentage for five selected areas per country. \cref{fig:protected_areas_v2} visually shows the encroachment. The results reveals that for certain protected areas in the WDPA up to 80 \% of the surface are covered by cocoa plantations. They also show a large difference in deforestation across protected area types that requires significantly more investigation. 

The most deforested protected areas in Côte d'Ivoire are the classified forests of Niegre, Scio and Mt. Sassandra, with 81.8 \%, 68.2 \% and 49.0 \% of their area under cocoa cultivation, respectively. All three of them have been exposed to illegal farming for decades \cite{barima2016cocoa, bitty2015cocoa}. Similarly, forest reserves such as Tano Ehuro, Manzan and Upper Wassaw in Ghana have severe forest clearing \cite{owubah2000forest, gyamfi2021insights} with cocoa expansion occurring within 23 \% to 77 \% of their surface. These high levels of deforestation in protected areas confirm and extend what has been found in Abu et al. for a very small subset of protected areas. 
For protected areas "of highest protection" in Ghana (e.g. Kakum National Park) our map detects almost no illegal cocoa plantations (1.0 \%). However, some of the national parks in Côte d'Ivoire are highly affected by illegal cocoa farming. In line with recent literature \cite{earth2017chocolate, wcf_annual_report, denis2015parc}, we are able to quantify the spatial extent of cocoa plantations within the protected areas such as over 6,400 ha and 2,700 ha in Mt. Péko National Park and Marahoué National Park, respectively. Yet, Tai National Park, a World Heritage Site and one of the largest protected areas in Côte d'Ivoire with  has experienced very little deforestation for cocoa. 

Part of the reason for the high overall deforestation rate - aside the many underlying drivers outlined in the discussion - is that some of the protected areas in the WDPA had already been degazetted early in the study period, thereby allowing cocoa production. In other regions particularly in Ghana, some villages and farms known as "admitted communities and/or farms" are legally allowed to remain in the forest reserves and to farm within delineated boundaries. However, it is known that these rights have been misused to further expand into remaining forests \cite{acheampong2019deforestation}.

\begin{figure}[!tb]
    \centering
    \begin{subfigure}[b]{\textwidth}
        \caption{}\label{fig:ndvi:ndvi_per_district}
        \includegraphics[width=\textwidth, page=1]{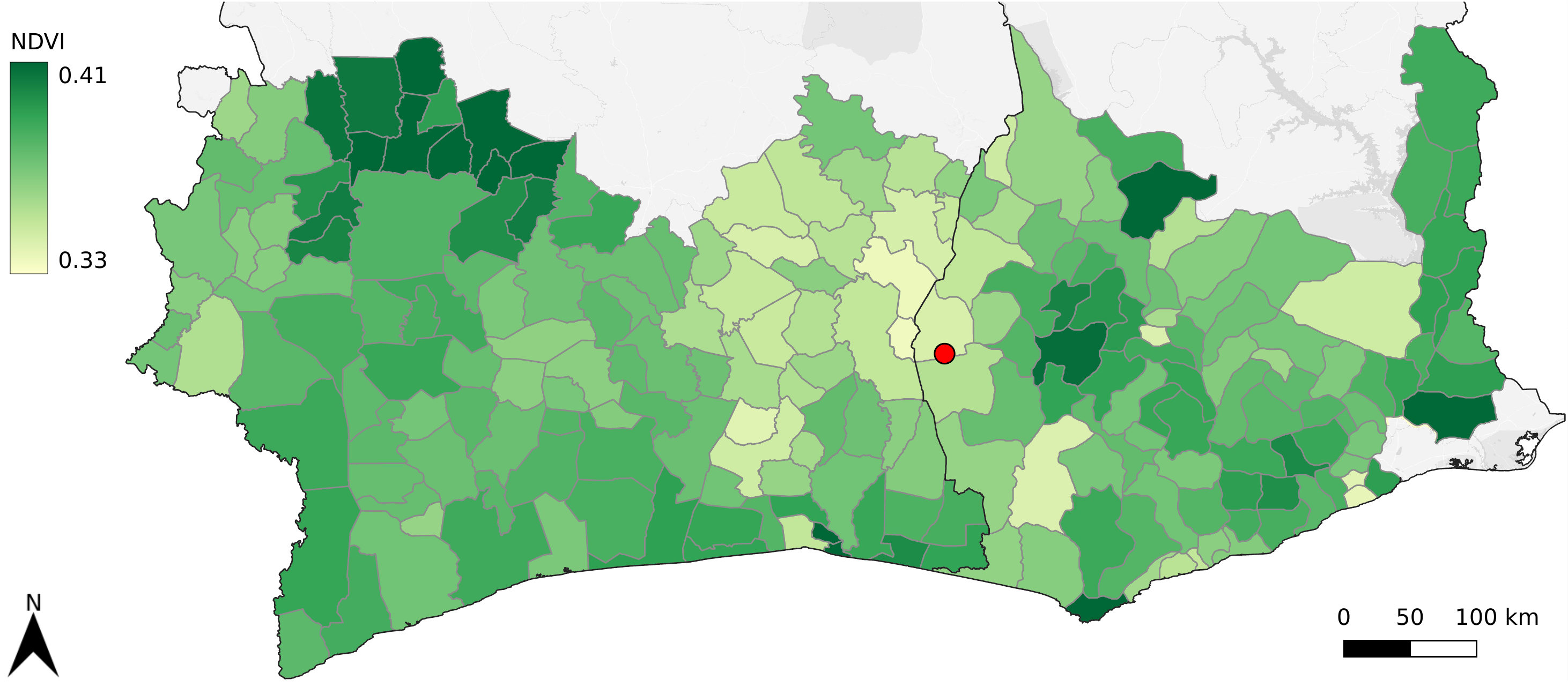}
    \end{subfigure}
    \begin{subfigure}[b]{\textwidth}
        \caption{}\label{fig:ndvi:ndvi_per_pixel}
        \includegraphics[width=\textwidth, page=2]{images/figure_5.pdf}
    \end{subfigure}
    \caption{\textbf{Vegetation health of cocoa, measured by normalized difference vegetation index (NDVI).} (a) Computing NDVI per district using only locations with actual cocoa plantings diminishes biases due to other crop types of vegetation and (b) allows targeted actions up to farm level when comparing at the native map resolution. The red dot in (a) indicates the approximate location in (b).}
\end{figure} 

Our map makes it possible to compute further vegetation parameters specifically for regions where cocoa is grown, while excluding other vegetation, i.e. plants that are not mapped as cocoa by our model. Naturally, the computed values can be influenced by shade trees and other vegetation within agroforestry systems. This segregation of cocoa and other vegetation allows us to use normalized difference vegetation index (NDVI) to monitor cocoa health on a large scale, with analyses at either pixel or district level, and to find regions where resources could be best used to improve the conditions of cocoa plantations. To demonstrate this, we measure vegetation health in terms of the NDVI. That index is based directly on the absorption of photosynthetically active radiation by leaves, and the re-emission of near-infrared radiation with too low photon energy. The index is defined as 
\begin{equation}
    \text{NDVI} = \frac{\text{NIR} - \text{RED}}{\text{NIR} + \text{RED}},
\end{equation}
where NIR and RED are the spectral reflectances in the near-infrared and red spectral bands, respectively. Higher NDVI at similar leaf area corresponds to better health. We showcase a fine-grained analysis at the district level for Côte d'Ivoire and Ghana, with the potential to identify local farming practices that promote the health of cocoa plants. In \cref{fig:ndvi:ndvi_per_district}, we depict the average NDVI per district for the time interval October 2018 to December 2021. The values range from slightly below 0.34 to slightly above 0.41, with an average of 0.38. There are clear regional differences. Particularly the border region between the Northeast of Côte d'Ivoire and the West of Ghana exhibits a cluster of lower NDVI values. We note that in Ghana the Western region is the principal cocoa growing area, with over one million hectares of cocoa. Yet, the low NDVI suggests worse plant health and lower productivity compared to, for instance, the Ashanti and Central regions. This could be further investigated, as the NDVI values could also be influenced by shade trees within agroforestry systems or local climatic characteristics. While NDVI at the district level is appealing in order to quickly identify larger areas that may have to be further investigated, it simultaneously becomes less expressive due to averaging effects over large areas. However, our map allows for fine-grained analysis up to farm or even sub-hectare level. As seen in \cref{fig:ndvi:ndvi_per_pixel}, we can compute vegetation health indices per pixel and year as an average over three months to reduce measurement noise in the corresponding Sentinel-2 bands. These maps can be used individually to compare the indices of neighboring areas or combined to produce annual difference maps. The former provides important information that may be directly relevant for farmers to improve the plants' health during the year and to monitor the impact of different weather patterns; while the latter may be an indicator of long-term developments, e.g. depleted soil.

\section*{Discussion}

In the following, we discuss benefits, implications and applications of our end-to-end framework and its product, concluding with the potential of our map to increase sustainability along the cocoa supply chain. 

Compared to existing mapping efforts, our framework promises a number of advantages. First, utilizing model ensembles in combination with aggregating over multiple satellite images of the same location allows for a confidence map in contrast to binary predictions. Consequently, an end-user gains an additional degree of freedom when using our map. Depending on the concrete tasks, one can adapt the threshold of classifying cocoa according to their needs, e.g. to fine-tune on a local region. Furthermore, the confidence map serves as guidance for measurements and improving predictions. While scores on the lower and upper range can be used to accurately and confidently take plant-specific measurements by decreasing the bias of false-positives, uncertain predictions can be preferably checked on the ground to improve the model performance. Hence, non-profit organizations, initiatives and governments can drastically reduce human resources for on ground surveys and mapping efforts. Mapping and protection efforts can be concentrated around and within protected areas. In case of reduced manpower, inspections on the ground can be focused around highly certain predictions. Additional to short-term forest clearance, it is possible to correlate long-term primary forest loss with cocoa encroachment in a highly accurate manner. 
We additionally tuned the confidence threshold on country-wide validation data. As demonstrated in the previous section, this results in a highly accurate binary cocoa map that can be used in various downstream tasks, e.g., to specifically mask out vegetation areas not used for cocoa production, in order to compute cocoa-specific vegetation indices such as the NDVI at local community level, thanks to the map's high resolution of 10 m. 

A major finding of our mapping efforts is the significant difference between the official harvested area in Ghana (1.63~Mha, average 2017--2020) and our total planted area estimation (2.7~Mha). Various reasons can partially explain this difference - which seems too important to only be due to young cocoa plantations detected by our map but not yet productive, and thus not accounted for in the harvested area. First, FAOSTAT numbers are based on imputed data, which are not as trustworthy for permanent crops due to unreliability of figures reported by the corresponding country, in particular for cocoa and coffee \cite{agricultural_production_fao}. In addition, it is known that up to 100,000 tonnes of cocoa beans per year have been smuggled across the border between the two countries back and forth \cite{cocoasmugglingafricabusiness, cocoasmugglinggurdian}, resulting in skewed official production figures.
Our map also differs from the official Ghanaian maps \cite{ghana_national_land_use_map}, which are only partially mapping cocoa in the country. In particular, the Volta region is ignored completely even though official production numbers (as seen in Section "Production") suggest cocoa farming takes place in the eastern part of the country. Our mapping efforts predicted a total of 60,000 ha additional cocoa plantations in the Volta region.

Our map demonstrates the massive role of cocoa in forest clearance in protected areas. Cocoa-driven deforestation is rooted in many interrelated factors, extensively explored in other
studies \cite{bymolt2018demystifying, barima2016cocoa, bitty2015cocoa}. At a basic level farmers pursue cocoa in protected activities to meet essential livelihood needs, including income and food production, which have been disrupted due to declining productivity, civil unrest \cite{barima2016cocoa}, migration and population pressure in existing farming areas. In both countries, nearly all cocoa smallholders live under the poverty line, with daily wages of less than one dollar a day \cite{earth2017chocolate}. Average cocoa yields are low, mainly due to depleted soils, ageing and diseased trees and low input use \cite{tio2017land}. Concomitant to securing land rights, clearing natural forests to establish new cocoa farms provides farmers with temporarily fertile land and thus higher yields and income in the short-run, compared to already cultivated sites\cite{ruf2015climate}. Our findings stress the drastic need for fairer prices and improved government and company policies to support cocoa farmers' adoption of improved practices. This must happen alongside stronger law enforcement to avoid rebound effects and preserve the remaining forests of Côte d'Ivoire and Ghana. 

We have developed an end-to-end trainable framework to map cocoa in the world's largest cocoa producing countries, promising high accuracy and flexibility. Nevertheless, a few limitations have to be considered. We have demonstrated the applicability and usefulness of deep learning for automated crop identification with optical satellite imagery. Yet, the final map relies on multiple image acquisitions for each location to cope with atmospheric disturbances, as optical sensors are limited by cloud cover. Due to this limitation, while the map itself can be used to detect cocoa within protected areas, it is not yet possible to capture new cocoa plantations on a weekly or monthly update. Integrating radar-based observations, in particular synthetic aperture radar (SAR) as an additional input for our framework could likely reduce the number of images needed per location, ultimately increasing the update rate of our map. Combining historical satellite data with our map to detect past and current cocoa expansion rates is also an interesting future application. The proposed approach is a generic framework not limited to a specific region and is expected to generalize to new areas. Given reference data from new regions of interest, the model may be fine-tuned to adapt to local conditions and patterns. In particular, regions with similar landscape characteristics (e.g., Cameroon or Nigeria) should only need small additional datasets, whereas adapting to countries such as Malaysia, Indonesia or Honduras with challenging growing practices, e.g., high shade tree density or mixed cultivation, will likely require a lot more reference data.

We believe that our study vindicates automatic analysis of satellite imagery as a tool for large-scale mapping of cocoa, and thereby presents a step forward in analyzing the cocoa supply chain and its sustainability implications. Beyond the cocoa supply chain this study also highlights the potential of using satellite imagery to derive the spatial extent of agricultural production in contexts with limited land documentation and therefore opens up opportunities to inform the design and implementation of public and private sustainability initiatives.  

\section*{Methods}

\subsection*{Data}

As model input, we use publicly available optical satellite imagery to train and apply our predictor. Data for both countries are acquired from the Copernicus Sentinel-2 mission. Sentinel-2 images consist of thirteen spectral bands, ranging from short-wave infrared to visible at a resolution of at most ten meters. We discard bands with a resolution of 60 meters and bilinearly upsample all 20-meter spectral bands to 10 meter resolution, for a total of nine channels that serve as input to our neural network. To make the model more robust towards atmospheric noise, cloudy recordings and sensor noise, we obtain in each Sentinel-2 grid tile the ten images with the lowest cloud cover within each 6-month period, over a total observation window of three years between October 2018 and December 2021. Consequently, for each pixel location, we have up to 60 valid observations. In most cases the number is smaller due to local cloud cover in some of the images (according to the Sentinel-2 basic cloud mask). Both during training and testing, such cloudy samples are masked with \emph{nodata} values in a post-processing step.

We obtained ground truth from different data providers, mainly industrial partners, cocoa foundations and nonprofit organizations. In total, we collected over 100,000 GPS-tracked cocoa farms and manually labelled over 10,000 background polygons of different sizes across both Ghana and the Ivory Coast. These negative samples, needed for supervised end-to-end learning, consist of villages, cities, rivers, lakes, open land, shrub land, forest and different commodities such as oil palm, rice or rubber. While cocoa has the largest number of individual polygons, the background samples have more than four times larger surface area due to the small size of cocoa farms (see Section "Deep learning framework" on how we account for this class imbalance during training). Instead of splitting our dataset into training and validation parts at the farm level, we randomly crop out large connected regions as validation areas, so as to avoid biases caused by spatial correlation between nearby farms \cite{ploton2020spatial}. In total we hold out $\approx\,$20\% as validation data. For training, we project and rasterize every ground truth polygon to the corresponding Sentinel-2 tile, randomly choose a patch of 320 $\times$ 320 meters (32 $\times$ 32 pixels) in which at least 10\% of the pixels are labelled, and extract the corresponding patch from a randomly selected Sentinel-2 image. That procedure is repeated to generate hundreds of millions of input samples. The statistical strength afforded by this massive amount of training data is one reason for the good performance of our framework. 

\subsection*{In situ data}

Independent from training and validation data, we create a unique evaluation protocol by gathering additional test data on the ground. Together with our partners from industry, we designed a verification campaign by choosing over 2,000 random locations around ten different cities in Côte d'Ivoire, in such a way that they overlap neither with the training nor with the validation set. Each location was defined by a center coordinate and an area around that center point, which may vary in size and shape. Several teams were sent out to visit these pre-defined sites. Whenever possible, they were instructed to walk around the boundary of the area and to report back the estimated percentage of cocoa grown on the site. Furthermore, they were asked to note down any other commodity grown within the area, i.e.,  an exemplary feedback would be that the area was occupied by "60 \% cocoa, 20 \% natural forest, 10 \% manioc, 10 \% palm trees". If the majority (\textgreater50 \% of the total area) was cocoa, the location was considered as a positive cocoa sample, otherwise a negative (non-cocoa) one. The on-site verification lasted for more than six months, beginning in March 2021. In that dataset the actual cocoa plantings are not geo-located within the polygons (smallholders in general grow multiple crops on their territory), thus we evaluate our map on the farm level and count a polygon as cocoa if a majority of its pixels falls into that class.

\subsection*{Deep learning framework}

With the aforementioned database, we were able to train end-to-end from raw spectral values, bypassing the manual design of predictive features. Instead, feature extraction is learned automatically. To that end, we employ a fully convolutional neural network based on the architecture proposed by Lang \etal~\cite{lang2019country}. The entry block of our model receives an image patch with nine Sentinel-2 bands and passes it through three consecutive residual blocks \cite{he2016deep} with learnable 1 $\times$ 1 convolutional filters. The output of that purely spectral, per-pixel analysis is then fed into a series of six residual blocks with 3 $\times$ 3 depth-wise separable convolutional layers \cite{chollet2017xception}, which enable the network to exploit textural features (i.e., spatial correlations between pixels). Next the (normalized) vegetation height map is included, by simply adding it to every channel of the intermediate feature map, and the result is fed through two further separable residual blocks to obtain the final feature representation. From it, the final output is computed with a single convolutional layer with two 1 $\times$ 1 filters, whose 2-channel output is passed through a Sigmoid transformation. This yields, at every 10 $\times$ 10 m pixel, two positive output values that sum to 1 and can be interpreted as the probabilities for the presence, respectively absence of cocoa. Since there are no downsampling operations and padding is applied in all residual blocks, the input resolution is retained and one can directly compare the output to the ground truth map. Additionally, as the network architecture is fully convolutional, it is not fixed to a specific spatial input size and can process image tiles of any size (subject to computing memory) during inference, reducing computation time during deployment. 

We optimize the neural network's weights by minimizing the Dice coefficient, also called the overlap index. The dice coefficient loss is defined as
\begin{equation}
    \mathcal{L} = \sum\limits_c \bigg(1 - \frac{2\sum\limits_i p_{ci}g_{ci} + \epsilon}{\sum\limits_i p_{ci} + \sum\limits_i g_{ci} + \epsilon}\bigg),
\end{equation}
where $c$ is the number of classes, $i$ the pixel index, $p$ and $g$ the prediction and ground truth, respectively. For numerical stability, a small $\epsilon$ is added to the numerator and denominator. The Dice loss is a common loss function used in medical image segmentation, as it is more robust under data imbalance compared to loss functions based on standard cross-entropy \cite{wang2018focal, milletari2016v}. As our training data is sparse within patches (a training patch only needs to have ground truth at \textgreater10 \% of its pixels), we further mask out all pixels without a ground truth label and compute the Dice loss selectively only for the labelled part. Patches are combined into batches of size 32. The network is trained for 500 epochs, with each epoch consisting of 40,000 iterations,
using the Adam optimizer \cite{kingma2014adam} with a base learning rate of $10^{-5}$. On our high-performance computing infrastructure, one training run took slightly more than 5 days. 

Confidence (respectively, uncertainty) estimates from individual deep neural networks are known to be poorly calibrated \cite{guo2017calibration}. For better uncertainty calibration we employ a model ensemble \cite{lakshminarayanan2017simple}. Ten replicas of the neural network just described are trained independently on the same dataset, with different random initializations and different (random) batches. Additionally, averaging estimates over multiple observations diminishes the influence of faulty classifications due to noisy observations.

\subsection*{Vegetation height map}

Besides the nine Sentinel-2 optical bands, our network ingests a dense vegetation height map as an auxiliary input channel (see previous section). The per-pixel vegetation heights have been derived from Sentinel-2 optical images with a deep learning method originally developed and tested for Southeast Asia \cite{lang2021high}. 
That method also employs a fully convolutional neural network, but which is trained to regress canopy height from Sentinel-2 imagery, using as training target sparse canopy height samples extracted from NASA's GEDI mission \cite{lang2022global}.
Due to the sparsity of GEDI's LiDAR footprints, we train this model not only on samples from Côte d'Ivoire and Ghana, but on an extended training area covering entire West Africa. Despite being trained on sparse data, the model outputs a dense canopy height map with 10 m ground sampling distance. 
Although the vegetation height map is also derived from Sentinel-2, and therefore arguably just another feature that could be learned from the input imagery, there are two reasons to directly incorporate it as an input channel. On the one hand, cocoa trees are known to only grow to a maximum height of $\approx$8 meters \cite{gardens2013plant, blaser2021effectiveness} (sometimes under higher shade trees, but these scattered trees protruding from the lower cocoa plants also provide a distinctive height pattern). Hence, vegetation height is an obvious predictive feature simply for its ability to identify high vegetation as not being cocoa. It therefore seems reasonable to simplify the learning process and save model capacity, by supplying it directly.
On the other hand, there is a more essential reason why we expect the separate tree height estimator to improve the estimates, namely that it brings in additional information. While the vegetation height map is indeed based on the same \emph{input}, it has not been learned from the same \emph{output}. Rather, the cocoa mapping pipeline benefits from the additional, strong supervision signal of the GEDI LiDAR measurements, which is baked into the canopy height map.

\section*{Data availability}

The cocoa probability map and its thresholded version will be released for download and will be available in the Google Earth Engine.
Both maps can be explored interactively in the following Google Earth Engine application:

\noindent\href{https://nk.users.earthengine.app/view/cocoa-map}{https://nk.users.earthengine.app/view/cocoa-map}.

\section*{Code availability}
Code is available at \url{https://github.com/D1noFuzi/cocoamapping/}. 

\section*{Acknowledgements}

The project received funding from Barry Callebaut Sourcing AG, as part of a Research Project Agreement. In particular, we thank Barry Callebaut Sourcing AG for realizing the ground campaign together. This research was funded through the 2019-2020 BiodivERsA joint call for research proposals, under the BiodivClim ERA-Net COFUND programme, and with the funding organisations Swiss National Science Foundation. We greatly appreciate the open data policies of the ESA Copernicus program.

\bibliography{sample}

\clearpage
\section*{Supplementary information}


\begin{table}[ht]
\centering
\begin{tabularx}{1\textwidth}{Y|Y|Y|Y}
\hline
Country & Region & Derived planted area (Mha) & Land cover (\%)\\
\hline
\multirow{6}{*}{Ghana} & Western region & 1.09 & 44.6 \\
& Eastern region & 0.29 & 15.5 \\
& Central region & 0.32 & 33.0 \\
& Volta region & 0.06 & 2.7 \\
& Brong Ahafro region & 0.29 & 7.3 \\
& Ashanti region & 0.66 & 26.7 \\
\hline
\multirow{10}{*}{Côte d'Ivoire} & Yamoussoukro region & 0.01 & 6.2 \\
& Woroba region & 0.03 & 1.0 \\
& Montagnes region & 0.95 & 31.2 \\
& Sassandra-Marahoué region & 0.54 & 22.3 \\
& Lacs region & 0.16 & 6.0 \\
& Lagunes region & 0.47 & 23.8 \\
& Gôh-Djiboua region & 0.60 & 38.5 \\
& Bas-Sassandra region & 1.18 & 42.9 \\
& Comoé region & 0.36 & 24.7 \\
& Zanzan region & 0.03 & 0.9 \\
\hline
\end{tabularx}
\caption{Total planted area and relative land cover per region in Côte d'Ivoire and Ghana. Regions are defined following official administrative boundaries.}\label{tab:districts}
\end{table}

\def\arraystretch{1.0}
\begin{table*}[!tb]
    \small
    \centering
    \begin{tabularx}{1\textwidth}{p{3cm}|Y|Y||p{3cm}|Y|Y}
    \hline
    \multicolumn{3}{l||}{Côte d'Ivoire} & \multicolumn{3}{l}{Ghana}\\
    \hline
    Protected area & Cocoa (ha) & Land cover (\%) & Protected area & Cocoa (ha) & Land cover (\%)\\
    \hline\hline
Taï National. (NP) & 1,552 & 0.3 & Subri River (FR) & 726 & 1.2 \\
Rapide Grah (CF) & 96,089 & 60.1 & Bia Tawya (FR) & 41,727 & 77.1 \\
Haute Dodo (CF) & 75,281 & 51.3 & Tain Tributa. (FR) & 132 & 0.2 \\
Scio (CF) & 90,418 & 68.2 & Krokosua Hil. (FR) & 5,099 & 10.8 \\
Niegre (CF) & 108,256 & 81.8 & Anlo-Keta la. (RSWoII) & 0 & 0.0 \\
Seguela (CF) & 25,564 & 22.3 & Tano Ofin (FR) & 858 & 2.1 \\
Mt. Sassandr. (CF) & 54,946 & 49.0 & Ankasa (RR) & 776 & 2.1 \\
CFNU No.72 (CF) & 51,080 & 48.7 & Kyabobo Nati. (NP) & 6 & 0.0 \\
Marahoue Nat. (NP) & 8,352 & 8.1 & Kogyae (SNR) & 0 & 0.0 \\
Mont Sangbe . (NP) & 0 & 0.0 & Bia North (FR) & 1,010 & 2.8 \\
Songan/Tamin (CF) & 35,763 & 39.0 & Subuma (FR) & 997 & 2.8 \\
Mabi/Yaya (CF) & 16,299 & 18.2 & Sui River (FR) & 3,497 & 9.8 \\
CFNU No.77 (CF) & 8,730 & 9.8 & Mpameso (FR) & 741 & 2.2 \\
CFNU No.58 (CF) & 62,359 & 72.3 & Bia National. (NP) & 891 & 2.8 \\
Go Bodienou (CF) & 39,275 & 47.9 & Bia Resource. (RR) & 723 & 2.3 \\
Yarani (CF) & 2,720 & 3.7 & Boin River (FR) & 764 & 2.5 \\
Issia (CF) & 18,603 & 27.9 & Bia National. (UBR) & 18 & 0.1 \\
Nibi Hana (CF) & 29,982 & 48.3 & Manzan (FR) & 15,512 & 56.1 \\
Duekoue (CF) & 18,096 & 29.7 & Asenanyo (FR) & 2,232 & 8.6 \\
Sangoue (CF) & 21,373 & 36.1 & Asukese (FR) & 87 & 0.3 \\
Bayota (CF) & 18,204 & 31.5 & Yoyo (FR) & 270 & 1.2 \\
Mt. Ko (CF) & 5,295 & 9.2 & Bomfoun (FR) & 0 & 0.0 \\
Dogodou (CF) & 21,308 & 39.9 & Subin (FR) & 533 & 2.3 \\
Irobo (CF) & 13,225 & 25.3 & Draw River (FR) & 193 & 0.9 \\
De (CF) & 16,386 & 31.5 & Atewa Range (FR) & 210 & 1.0 \\
Tene (CF) & 6,763 & 14.5 & Tano Ehuro (FR) & 16,275 & 77.6 \\
Besso (CF) & 19,732 & 44.2 & Kakum (NP) & 256 & 1.2 \\
Beki Bosse M. (CF) & 12,096 & 28.8 & Tano Nimri (FR) & 669 & 3.3 \\
Diambarakrou (CF) & 26,581 & 64.2 & Pamu Berekum (FR) & 89 & 0.5 \\
CFNU No.43 (CF) & 12,993 & 32.1 & Bia Tano (FR) & 195 & 1.0 \\
CFNU No.66 (CF) & 1,141 & 3.0 & Afram Headwa. (FR) & 60 & 0.3 \\
Plaine des E. (CF) & 7,489 & 20.2 & Bodi (FR) & 12,947 & 70.0 \\
Kamesso (CF) & 10 & 0.0 & Chai River (FR) & 22 & 0.1 \\
Abeanou (CF) & 4,550 & 13.6 & Bandai Hills. (FR) & 0 & 0.0 \\
Tanoe (CF) & 18,332 & 55.7 & Fure Headwat. (FR) & 669 & 4.0 \\
Monogaga (CF) & 20,937 & 65.3 & Oda River (FR) & 306 & 1.8 \\
Gouin (CF) & 19,099 & 61.2 & Bonsa Ben (FR) & 178 & 1.1 \\
Mont Peko Na. (NP) & 6,479 & 21.5 & Bonsa River (FR) & 1,059 & 6.5 \\
Seguie (CF) & 8,850 & 29.9 & Worobong Nor. (FR) & 1,852 & 11.3 \\
CFNU No.50 (CF) & 2,862 & 9.7 & Dadieso (FR) & 303 & 1.9 \\
CFNU No.30 (CF) & 3,552 & 12.2 & Fure River (FR) & 161 & 1.0 \\
Yalo (CF) & 2,941 & 10.1 & Assin Attand. (RR) & 387 & 2.5 \\
CFNU No.40 (CF) & 2,760 & 9.8 & Desiri (FR) & 9,383 & 60.2 \\
N'Zo Fauna R. (PFR) & 13 & 0.0 & Boabeng-Fiem. (WS) & 0 & 0.0 \\
Mopri (CF) & 10,742 & 39.6 & Togo Plateau (FR) & 301 & 2.0 \\
Arrah (CF) & 7,809 & 29.6 & Tano Anwia (FR) & 212 & 1.4 \\
CFNU No.56 (CF) & 12,007 & 47.2 & Tonton (FR) & 161 & 1.1 \\
CFNU No.31 (CF) & 228 & 0.9 & Bosomkese (FR) & 250 & 1.7 \\
Tiapleu (CF) & 5,416 & 22.0 & Sukusuki (FR) & 12,053 & 83.6 \\
Mt. Tia (CF) & 14,886 & 62.0 & Southern Sca. (FR) & 1,486 & 10.7 \\
Ebrinenou (CF) & 12,744 & 53.6 & Upper Wassaw (FR) & 3,198 & 23.6 \\
Moyenne Mara. (CF) & 1,280 & 5.7 & Bonsam Bepo (FR) & 807 & 6.1 \\
Azagny Natio. (NP) & 249 & 1.1 & Nini-Suhien (NP) & 665 & 5.1 \\
    \end{tabularx}
\end{table*}
\begin{table}[!tb]
    \small
    \centering
    \begin{tabularx}{1\textwidth}{p{3cm}|Y|Y||p{3cm}|Y|Y}
    \hline
    \multicolumn{3}{l||}{Côte d'Ivoire} & \multicolumn{3}{l}{Ghana}\\
    \hline
    Protected area & Cocoa (ha) & Land cover (\%) & Protected area & Cocoa (ha) & Land cover (\%)\\
    \hline\hline
CFNU No.53 (CF) & 2,619 & 12.5 & Awura (FR) & 0 & 0.0 \\
Agbo (CF) & 12,029 & 57.8 & Neung South (FR) & 399 & 3.1 \\
Zuoke (CF) & 10,825 & 52.1 & Pra Anum (FR) & 984 & 7.6 \\
Abokouamekro. (NR) & 4 & 0.0 & Opon Mansi (FR) & 646 & 5.0 \\
Niouniourou (CF) & 12,375 & 64.5 & Opro River (FR) & 19 & 0.2 \\
Hein (CF) & 2,546 & 13.9 & Bosomoa (FR) & 0 & 0.0 \\
Tyemba (CF) & 0 & 0.0 & Esukawkaw (FR) & 124 & 1.0 \\
Zagoreta (CF) & 8,844 & 50.0 & Southern Sca. (FR) & 277 & 2.3 \\
Vavoua (CF) & 4,186 & 23.8 & Anhwiaso Eas. (FR) & 1,033 & 8.5 \\
Brassue (CF) & 5,034 & 28.8 & Boi Tano (FR) & 210 & 1.7 \\
Semien Flans. (CF) & 4,536 & 26.0 & Tinte Bepo (FR) & 316 & 2.6 \\
Elroukro (CF) & 2,661 & 15.3 & Ayum (FR) & 134 & 1.1 \\
Adzope (CF) & 6,500 & 38.3 & Worobong Sou. (FR) & 104 & 0.9 \\
Tos (CF) & 1,477 & 8.9 & Bowiye Range (FR) & 222 & 2.0 \\
Kavi (CF) & 9,125 & 57.8 & Chirimfa (FR) & 3 & 0.0 \\
Oume Doka (CF) & 4,213 & 27.0 & Bimpong (FR) & 233 & 2.2 \\
Marahoue (CF) & 2,789 & 18.0 & Pra Suhyien . (FR) & 373 & 3.6 \\
Tankesse (CF) & 2,307 & 15.3 & Bura River (FR) & 113 & 1.1 \\
Kravassou (CF) & 5,468 & 37.2 & Nkrabia (FR) & 244 & 2.4 \\
Ira (CF) & 5,871 & 41.1 & Tano Suhyien (FR) & 407 & 4.5 \\
Manzan (CF) & 5,596 & 39.9 & Mankrang (FR) & 1 & 0.0 \\
Mt. De (CF) & 1,900 & 14.0 & Kwamisa (FR) & 234 & 2.8 \\
Bouafle (CF) & 2,163 & 16.1 & Fum Headwate. (FR) & 158 & 1.9 \\
Dassieko (CF) & 4,692 & 34.8 & Kabo River (FR) & 1,544 & 19.0 \\
CFNU No.39 (CF) & 7,484 & 55.9 & Dampia Range (FR) & 414 & 5.1 \\
Kassa (CF) & 2,617 & 19.9 & Dome River (FR) & 533 & 6.7 \\
Offumpo (CF) & 6,637 & 51.2 & Pra Suhyien . (FR) & 76 & 1.0 \\
Ananguie (CF) & 7,712 & 60.5 & Asuokoko Riv. (FR) & 157 & 2.0 \\
Nizoro (CF) & 7,477 & 59.4 & Afrensu Broh. (FR) & 0 & 0.0 \\
Mt. Sainte/C. (CF) & 7,612 & 62.7 & Tano Suraw E. (FR) & 2,216 & 28.6 \\
Bableu (CF) & 5,249 & 43.6 & Kalakpa (RR) & 0 & 0.0 \\
Krozalie (CF) & 5,486 & 48.7 & Asubima (FR) & 1 & 0.0 \\
Mt. Bolo (CF) & 4,929 & 44.0 & Bosumtwi Ran. (FR) & 1,089 & 14.1 \\
CFNU No.68 (CF) & 6,683 & 59.7 & Bomfobiri (WS) & 333 & 4.3 \\
Mando (CF) & 2,016 & 18.7 & North Bandai. (FR) & 0 & 0.0 \\
CFNU No.42 (CF) & 4,387 & 41.7 & Bonkoni (FR) & 53 & 0.7 \\
CFNU No.34 (CF) & 4,946 & 47.2 & Jema Asemkro. (FR) & 1,925 & 27.1 \\
Kouadikro (CF) & 246 & 2.4 & Ndumfri (FR) & 156 & 2.2 \\
CFNU No.36 (CF) & 96 & 0.9 & Apepesu Rive. (FR) & 1,084 & 16.2 \\
Bolo (CF) & 4,676 & 47.1 & Muro (FR) & 183 & 2.8 \\
CFNU No.74 (CF) & 1 & 0.0 & Sawsaw (FR) & 9 & 0.1 \\
CFNU No.33 (CF) & 2,514 & 27.3 & Mirasa Hills (FR) & 191 & 3.0 \\
Goudi (CF) & 4,548 & 50.2 & Totua Shelte. (FR) & 65 & 1.0 \\
CFNU No.57 (CF) & 589 & 6.5 & Northern Sca. (FR) & 0 & 0.0 \\
CFNU No.38 (CF) & 2,684 & 30.7 & Nsuensa (FR) & 10 & 0.2 \\
    \hline
    \end{tabularx}
    \caption{Extended table of the one hundred protected areas ordered according to their total protected area in Côte d'Ivoire and Ghana (within suitable cocoa growing area). CFNU = classified forest unknown name, CF = classified forest, NP = national park, WS = wildlife sanctuary, FR = forest reserve, RR = resource reserve.}\label{tab:protected_areas_extended}
\end{table}

\end{document}